\documentclass[conference]{IEEEtran}
\IEEEoverridecommandlockouts
\usepackage{cite}
\usepackage{amsmath,amssymb,amsfonts}
\usepackage{graphicx}
\usepackage{textcomp}
\usepackage{xcolor}
\usepackage{subfigure}
\usepackage{hyperref}

\newcommand{\li}[1]{{\color{black} #1}}

\usepackage[ruled,linesnumbered]{algorithm2e}
\SetKwComment{Comment}{/* }{ */}
\newcommand{\pluseq}{\mathrel{+}=}
\SetKw{KwBy}{by}

\def\BibTeX{{\rm B\kern-.05em{\sc i\kern-.025em b}\kern-.08em
    T\kern-.1667em\lower.7ex\hbox{E}\kern-.125emX}}
\begin{document}

\title{Inspector: Pixel-Based Automated Game Testing via Exploration, Detection, and Investigation   \\
\thanks{*Intern at MSRA.}
\thanks{†Corresponding author.}
}

\author{\IEEEauthorblockN{Guoqing Liu}
\IEEEauthorblockA{\textit{Microsoft Research Asia} \\
guoqingliu@microsoft.com}
\and
\IEEEauthorblockN{Mengzhang Cai*} 
\IEEEauthorblockA{\textit{University of Science and Technology of China} \\
caimz@mail.ustc.edu.cn}
\and
\IEEEauthorblockN{Li Zhao}
\IEEEauthorblockA{\textit{Microsoft Research Asia} \\
lizo@microsoft.com}
\and
\IEEEauthorblockN{Tao Qin†}
\IEEEauthorblockA{\textit{Microsoft Research Asia} \\
taoqin@microsoft.com}
\and
\IEEEauthorblockN{Adrian Brown}
\IEEEauthorblockA{\textit{Xbox Studios Quality} \\
adrianbr@microsoft.com}
\and
\IEEEauthorblockN{Jimmy Bischoff}
\IEEEauthorblockA{\textit{Xbox Studios Quality} \\
jimmyb@microsoft.com}
\and
\IEEEauthorblockN{Tie-Yan Liu}
\IEEEauthorblockA{\textit{Microsoft Research Asia} \\
tyliu@microsoft.com}
}

\maketitle

\begin{abstract}

Deep reinforcement learning~(DRL) has attracted much attention in automated game testing. 
Early attempts rely on game internal information for game space exploration, thus requiring deep integration with games, which is inconvenient for practical applications. 
In this work, we propose using only screenshots/pixels as input for automated game testing and build a general game testing agent, Inspector, that can be easily applied to different games without deep integration with games.
In addition to covering all game space for testing, our agent also tries to take human-like behaviors to interact with key objects in a game, since some bugs usually happen in player-object interactions.
Inspector is  based on purely pixel inputs and comprises three key modules: game space explorer, key object detector, and human-like object investigator. 
Game space explorer 
aims to explore the whole game space by using a curiosity-based reward function with pixel inputs. 
Key object detector aims to detect key objects in a game, based on a small number of labeled screenshots.
Human-like object investigator aims to mimic human behaviors for investigating key objects via imitation learning.
We conduct experiments on two popular video games: Shooter Game
and Action RPG Game\footnote{
from https://docs.unrealengine.com/4.27/Resources/SampleGames. }. 
Experiment results demonstrate the effectiveness of Inspector in exploring game space, detecting key objects, and investigating objects. Moreover, Inspector successfully discovers two potential bugs in those two games.
The demo video of Inspector is available at \url{https://github.com/Inspector-GameTesting/Inspector-GameTesting}.
\end{abstract}


\section{Introduction}
Game testing is a critical part of game development, and has been long recognized as a notoriously challenging task~\cite{b1, b2, b3, b4}.
As video games continue growing in both size and complexity, it has become more challenging to ensure that all the game content is well tested. 
Most game companies rely on human testers, or human written scripts for game testing, which is costly and time-consuming, especially for complex video games with large map sizes. As a result, many bugs are still undiscovered and left in games (e.g., Cyberpunk 2077) after the official release, and then discovered afterwards by game players, which seriously hurt players’ experiences. 

In recent years, reinforcement learning (RL) based automated game testing has attracted much attention in both academia \li{and} the game industry~\cite{b18,b19,b20,b21,b22}, due to its flexibility, low cost, and ease of scale up~\cite{b20}\cite{b22}\cite{b23}.  
While those methods have shown promising results in their tested games, they all rely on game internal state information such as player position, player velocity, game reward, etc., which has several limitations.  First, they may not \li{be} easily applied to other games, since the internal states of different games usually differ a lot, in terms of state dimensions, state structure, and update frequency. Second, to access game internal states, a testing program/service needs to be deeply integrated into game source codes and therefore deeply coupled with the game development process, which limits the application of those algorithms to only games with source code accessible.  Furthermore, even when the source code of a game is accessible, since the code is usually under fast change during the development stage, it is inconvenient for game developers always to maintain the deep integration of the testing code.

In this work, we build a general game testing agent/tool, named Inspector, purely based \li{on} game screenshots, i.e., our agent takes only pixel inputs to make decisions. Such a testing agent does not suffer from the above limitations: (1) It is internal state free and able to be applied to a variety of games with or without access to the source code of those games. In other words, such an agent can be provided as a general testing service for both first-party games (with source code available) and third-party games (with source code unavailable). (2) In principle, it can be easily applied to all video games, since we can simply resize the screenshots of different games to the same size.  
Another advantage of Inspector over previous methods is that in addition to covering all game space for testing, it also mimics human players’ behaviors to interact with key objects in a game, such as weapons, health packs, etc., as bugs usually happen in those player-object interactions~\cite{b39}. 

For these two purposes, Inspector comprises three components: a game space explorer, a key object detector, and a human-like investigator. (1) The explorer focuses on game space exploration, using a curiosity-based reward function which takes the screenshot as input. (2) The object detector focuses on key object detection and then the investigator takes human-like behaviors to interact with detected key objects. To reduce the cost of human labeling, we adopt a few-shot object detection method to effectively detect key objects based on a small number of labeled screenshots. The human-like investigator is trained via imitation learning from human players’ trajectories.  
As an integrated system, Inspector explores the game space; when it detects a key object, it investigates the object; after the investigation is done, it will continue exploring the game, and find the next object to investigate and test.

We conduct experiments on two video games: Shooter Game and Action RPG Game, from the two most popular game categories, shooter and action-adventure games, respectively. For game space exploration, we show that Inspector can achieve super-human coverage. For key object detection, we show that Inspector works well for most of the test cases, even for the cases in which the background is never seen in training examples. For human-like object investigation, we show that Inspector can well mimic human behaviors while interacting with key objects such as health packs. Moreover, Inspector also successfully discovers two potential bugs in the two games undiscovered before, which seems not easy to be found by human testers. 
We provide several demo videos in anonymous GitHub\footnote{\url{https://github.com/Inspector-GameTesting/Inspector-GameTesting}}.

To summarize, the main contributions of this paper are as follows:
\begin{enumerate}
    \item To the best of our knowledge, we are the first to design an automated game testing agent purely based on pixel inputs.  
    \item In addition to game space exploration, our agent can also detect key objects and take human-like behaviors to interact with the objects, so as to better expose hidden bugs.
   \item We conduct experiments on two popular video games and demonstrate the effectiveness of Inspector. 
\end{enumerate}

\section{Related Work}

In this section, we briefly review related work on game testing, with special focus on reinforcement learning for game testing.
\subsection {Game Testing}
As modern games continue growing both in size and complexity, game testing  has become more challenging; consequently, even the most popular games on the market lack sufficient testing~\cite{b1, b2, b3}. 
To improve efficiency and reduce cost,  game testing has gradually evolved from manual and ad-hoc approaches to automated testing~\cite{b5, b6, b7, b8, b9, b10, b11, b12,b13, b14, b15, b16, b17,b18,b19,b20, b21, b22, b23}. 
 \cite{b5} proposed black box testing and scenario-based testing for online games. They only need to refine the game description language and virtual game maps, instead of rewriting the virtual client dummy code. Complex scenarios such as attacks, party plays, and waypoint movements can be tested by combining actions.
\cite{b6} proposed a testing model specifically for the creation and execution of fully automated regression tests. The usability and veracity of records and playback techniques are combined with the possible test coverage of tests written in a game-specific scripting language.
However, when a game environment changes, test sequences and scenarios such as \cite{b5} and \cite{b6} become obsolete and human efforts are required to create new test sequences.
\cite{b7} proposed a model-based approach for automated test sequence generation and modeled platform games using domain modeling for representing the game structure and UML state machines for behavioral modeling. 
Unfortunately, generating sequences from UML will run into state explosions for large games. 
\cite{b8} designed an AI agent that plays according to a Petri net description of a game with high-level actions; a limitation is that it only generates test sequences that cover some specific game scenarios. 
\cite{b9} and \cite{b12} built MCTS~\cite{b41} based agents to exercise synthetic and human-like test goals. The sequences generated by the agents are replayed in the game to check for bugs by humans. 
 
\subsection{Reinforcement Learning for Game Testing}
 
The latest trend for game testing is to leverage reinforcement learning techniques~\cite{b26,b27,b28}, especially for maximizing the coverage of a game~\cite{b23}. 
\cite{b20} introduced a self-learning mechanism based on DRL to explore/exploit game mechanics with a user-defined reward signal.  
\cite{b21} leveraged DRL, evolutionary algorithms, and multi-objective optimization to perform automated game testing in two commercial combat games, which balances between winning a game and exploring the space of the game. 
\cite{b22} maximized game state coverage through DRL with a count-based reward function~\cite{b24}\cite{b25}. 
\cite{b23} leveraged human demonstrations to better cover the states of a game. 
All of the above works rely on game internal state information for game space coverage and thus a testing agent needs to access the source code of a game to collect such information. 
 
Such deep integration with game source code is inconvenient  when a game is under development with fast code changes, not to mention that the source code of most games are not available for a third-party testing tool. 
Different from these methods, our work uses only screenshot input for game testing, which is not limited by the availability of game source code and thus is easy to be applied to many games. Furthermore, our work enables the agent to interact with key objects of a game, so as to better expose hidden bugs.

\section{Inspector}



In this section, we introduce our pixel-based agent, Inspector, which serves two purposes for automated game testing: covering all game space, as did in most previous works, and interacting with key objects in games like human players/testers, which was ignored in most previous works. 
Inspector consists of three key modules: 
1) Game space explorer (Section~\ref{method:exploration}),
2) Key object detector (Section~\ref{method:detection}), 
3) Human-like object investigator (Section~\ref{method:investigation}). 
Putting them together brings us the integrated system of Inspector (Section~\ref{method:integrated_system}), which can not only explore all game space, but also detect key objects and investigate the objects in a human-like way. 
Combining Inspector with some existing works using deep learning methods to detect bugs from screenshots (e.g., \cite{b40}) leads to an end-to-end automated game testing service.  

\subsection{Game Space Explorer} \label{method:exploration}

\begin{figure}[t]
\centerline{\includegraphics[width=0.5\textwidth]{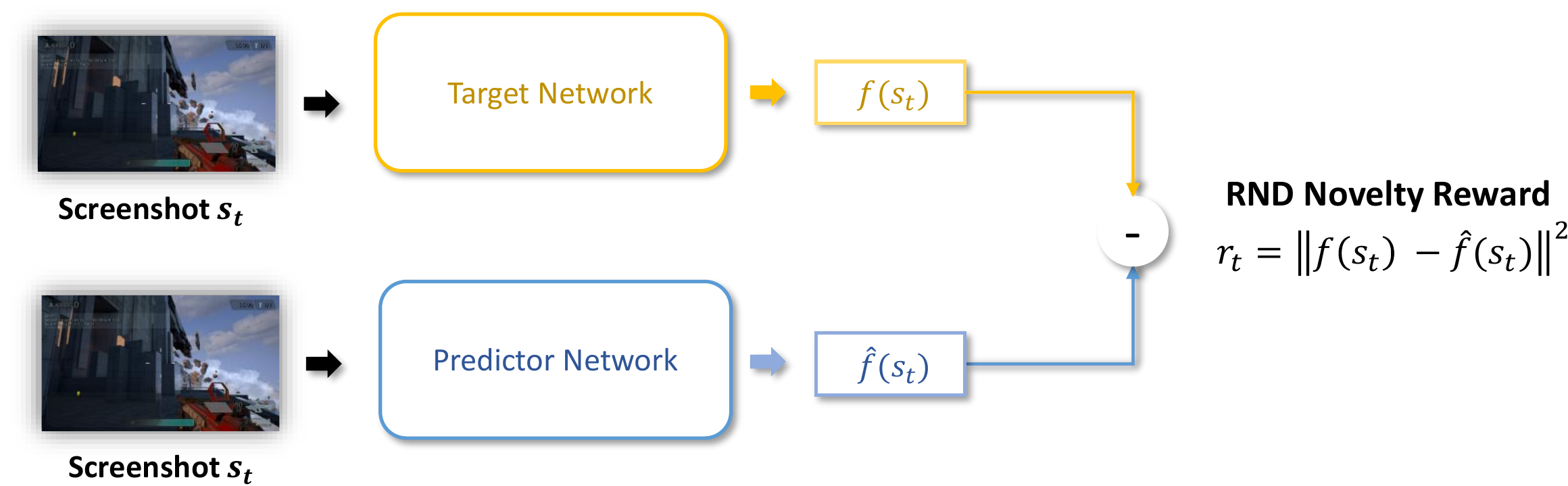}}
\caption{Illustration of RND-based curiosity reward for game space exploration.}
\label{fig:rnd}
\end{figure}

Exploring and covering all possible states of a game space is the basic requirement for automated game testing and well-studied in general reinforcement learning research (beyond game testing).  The key challenge here is how to design effective feedback signals to guide and encourage the exploration of new/unseen game states. Some recent work~\cite{b22} adopted the count-based reward function to encourage exploration, which counts the number of unique game internal states and uses this number to reward exploration. As mentioned before, we aim to design a general testing agent taking visual pixels as inputs rather than game internal states. However, it is hard to count high-dimensional, continuous pixels, and thus this kind of count-based reward function does not work for our setting. 
 
To address this challenge, we design a curiosity-driven reward function, based on Random Network Distillation~(RND)~\cite{b29}. 
Our RND-based reward function is inspired by an interesting property of deep neural networks (DNN): DNNs usually have lower prediction errors on those seen examples, while having relatively higher prediction errors on those unfamiliar examples. 
When we train a DNN on these seen screenshots during exploration,
the resulted prediction error is supposed to be large on those novel screenshots never seen before and vice versa. 
 
Specifically, two neural networks are introduced:  a randomly initialized and fixed target network $f: \mathcal{S} \to \mathbb{R}^{k}$ which sets the ground truth for prediction, 
and a predictor network  $\hat{f}: \mathcal{S} \to \mathbb{R}^{k}$ trained on the screenshots seen during exploration.
Both the target network and the predictor network map a screenshot to an embedding vector. 
The predictor network with parameter $\psi$ is trained to minimize the following mean square error:
\begin{equation}\label{eq:rnd_loss}
\psi^{*} = \min_{\psi \in \Psi} \frac{1}{N} \sum_{i=1}^{N}{||\hat{f}(s_i; \psi) - f(s_i)||^{2}},
\end{equation}
where $N$ denotes the number of screenshots the agent has ever seen during exploration.  
Then the reward function is defined as follows.
\begin{equation}\label{eq:rnd_reward}
r_{t} = ||f(s_t) - \hat{f}(s_t)||^{2}
\end{equation}
 
We illustrate this RND-based reward function in Fig.~\ref{fig:rnd}.
If a screenshot $s_t$ has been seen before, it will have a small prediction error and thus a low reward. 
With this reward function, the agent is motivated to explore novel and unseen screenshots.  We then use Proximal Policy Optimization~(PPO) algorithm~\cite{ b30, b31, b32}, which is widely used in various applications of DRL, to train the exploration policy in our game space explorer.

\subsection{Key Object Detector}
\label{method:detection}

\begin{figure}[t]
\centerline{\includegraphics[width=0.5\textwidth]{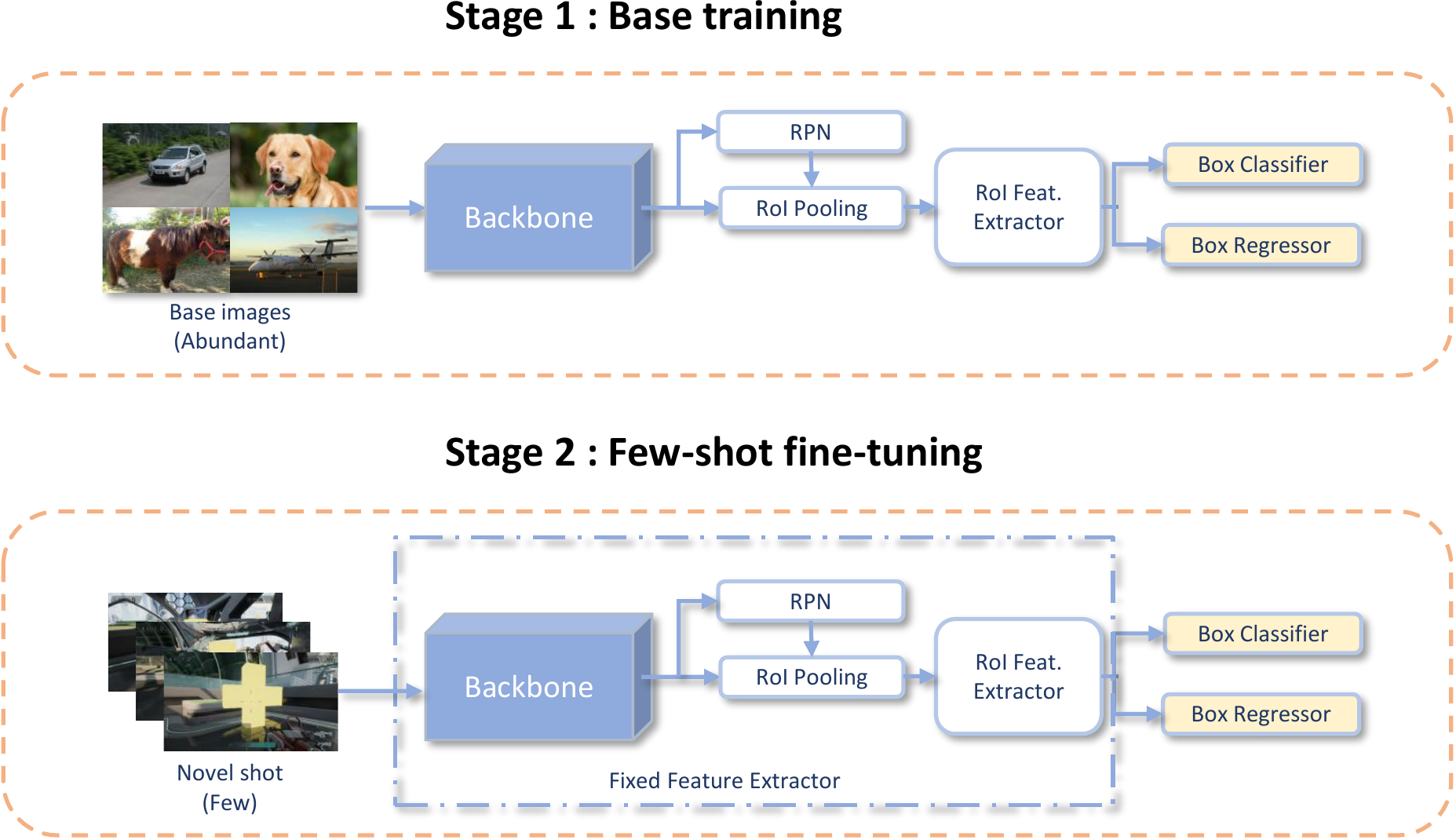}}
\caption{Illustration of 
two-stage fine-tuning method for key object detection based only on a few examples. 
}
\label{fig:detection}
\end{figure}
One important task for human testers is to interact with some key objects in a game, as some hidden and difficult-to-find bugs only show up when a player interacts with some objects~\cite{b39}. Therefore, different from most previous works which only focus on covering the whole game space, another focus of our agent is to  detect and investigate key objects in a way similar to human players/testers, to better expose this kind of difficult bugs.

In this subsection, we introduce the key object detector, which is to detect key objects in  a game during exploration.  
For example, in FPS games, the key object can be health packs, ammo, etc.
The technical challenge here is that it is costly and time-consuming to manually label a large number of screenshots for each kind of key object, especially for games with many kinds of key objects.
To reduce the cost of human labeling, we adopt a simple yet effective two-stage fine-tuning method for few-shot object detection~\cite{b33}. 
Specifically, we first pre-train the entire object detector using MS COCO dataset\footnote{publicly available in https://cocodataset.org.}, which is a large-scale dataset for general object detection.
Then, we only fine-tune the last layers
of the detector using a small number of labeled screenshots with each kind of key object, while freezing the other parameters of the model.
The illustration of the two-stage 
method is shown in Figure.~\ref{fig:detection}.

For the base detection model, we adopt the widely used Faster R-CNN~\cite{b34}, which consists of a feature extractor and a box predictor.
The feature extractor comprises the backbone (e.g., ResNet~\cite{b35}), the region proposal network (RPN), as well as a two-layer fully connected (FC) sub-network as a proposal-level feature extractor. 
The box predictor consists of a box classifier to classify the object categories and a box regressor to predict the bounding box coordinates.
Both the backbone features and the RPN features are class-agnostic and likely to transfer to novel classes~\cite{b33}. 
We fix the feature extractor and directly leverage these features learned from the base classes in the MS COCO dataset for the new class of key objects.

With the detector trained by the two-stage training with only a few samples, Inspector can recognize  key objects during exploration,  when an object is nearby, which then triggers the human-like investigator introduced in the following subsection. 
To determine if the object is nearby, we leverage the bounding box size from the detection model when the object has been detected.

\subsection{Human-like Object Investigator}
\label{method:investigation}

The goal of the human-like object investigator is to interact with key objects within a game in a human-like manner. Since different kinds of objects are usually interacted by human players in different ways, leverage imitation learning techniques~\cite{b36} to train an investigation policy for each kind of object. 
We first collect some demonstration trajectories for each kind of key object from human testers when they take actions to interact with and investigate the key object.
Then, we train an investigation policy (a convolutional neural network, CNN)  from human demonstrations by minimizing the behavior cloning (BC) loss~\cite{b36}:
\begin{equation}
\mathcal{L}_{BC}(\pi_{\theta}) =  - \sum_{(s, a) \in D}{\log \pi_{\theta}(a|s)},
\end{equation}
where $\pi_{\theta}$ denotes the learned policy, and $D$ denotes the set of (screenshot, action) pairs from demonstration trajectories. 

With the learned investigation policy, Inspector can  interact with and investigate  key objects to expose  potential bugs in a human-like manner.

\subsection{The Integrated System}
\label{method:integrated_system}



\begin{figure}[h]
\centerline{\includegraphics[width=0.45\textwidth]{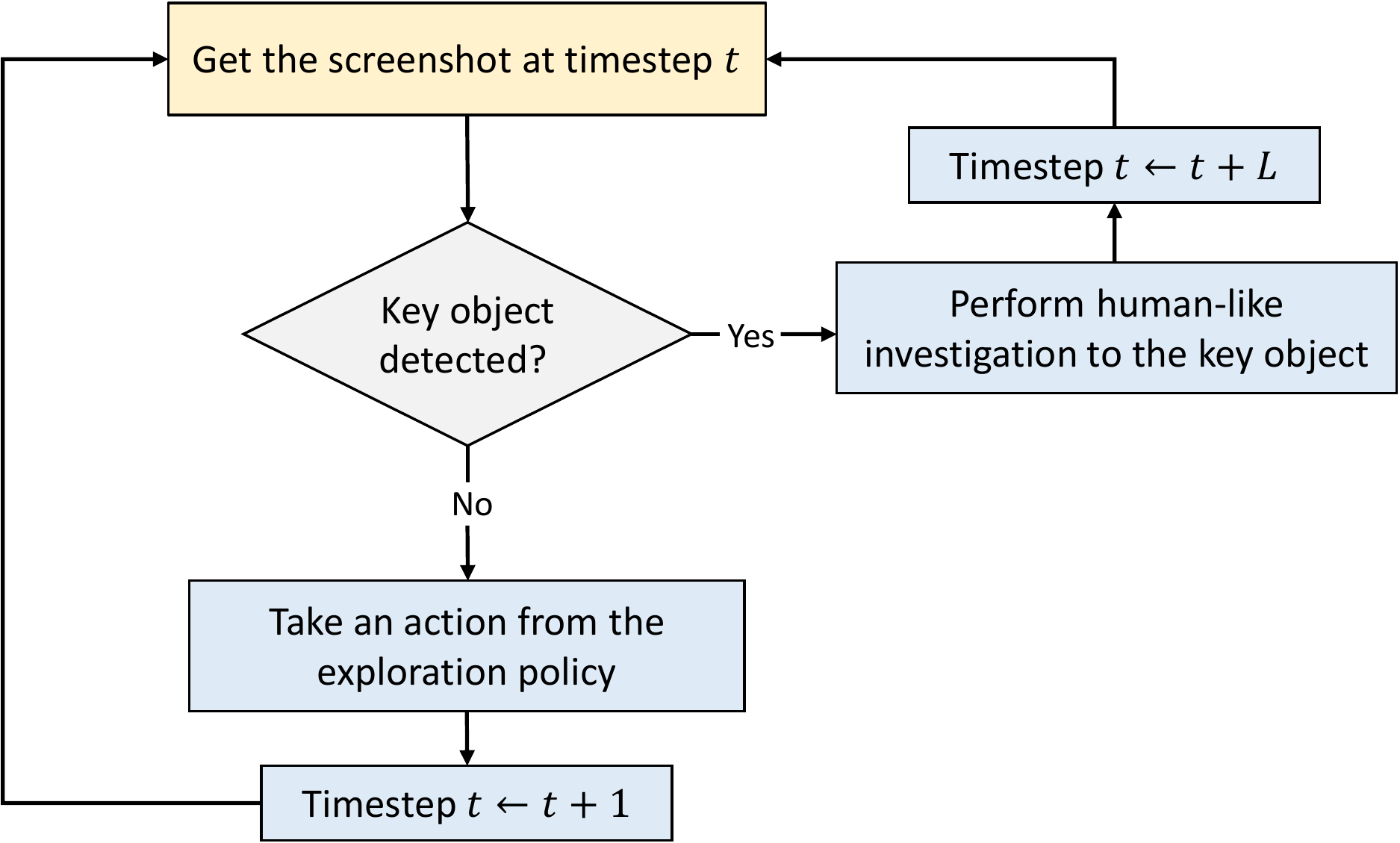}}
\caption{
The decision flow of the Inspector agent.
}
\label{fig:integrated_system}
\end{figure}

After introducing the three modules, in this subsection, we describe the integrated system, the Inspector agent.

The decision flow of the Inspector agent is shown in Fig.~\ref{fig:integrated_system}.
At each timestep $t$, the agent  first uses the detection models to detect whether there exists  a key object in the current screenshot.  
If not, the agent will take an action using the exploration policy to explore the game space; 
otherwise, the agent  starts to investigate the detected object by taking a series of actions from the investigation policy for that kind of object.
When the investigation is done, the agent will continue exploration, until it detects the next key object to investigate.
The pseudo-code for the integrated system is  provided in Algorithm~\ref{algo}.

\begin{algorithm}[h]\label{algo}
\caption{pseudo-code for the integrated system}
\textbf{Models:} the exploration policy $\pi_{\text{explore}}$, the fixed target network $f$, the predictor network $\hat{f}$, the trained key object detector $d$, the trained investigation policy $\pi_{\text{investigate}}$\;
$N \to$ number of timesteps\;
$L \to$  length of an investigation process\;
$M \to$ batch size for updating the parameters\;
$E\to $ number of optimization epochs\;
obtain the initial screenshot $s_0$\;
set current timestep $t=0$\;
\While{$t < N$}{
    use the key detection detector $d$ to infer current screenshot $s_t$\;
  \eIf{both the bounding box size and the classification probability exceed their thresholds}{
    investigate the key object using $\pi_{\text{investigate}}$ for $L$ timesteps\;
    $t \pluseq L$\;
  }
  {
  sample action $a_t \sim \pi_{\text{explore}}(a_t|s_t)$ and obtain the next screenshot $s_{t+1}$\;
  calculate the RND-based curiosity reward $r_{t}$\;
  add $s_t, a_t, s_{t+1}, r_{t}$ into optimization batch $\mathcal{B}$\;
  $t \pluseq 1$\;
   }
   \If{$t\mod M == 0$}
   {
  \For{$i\gets0$ \KwTo $E$}{
    optimize $\pi_{\text{explore}}$ with PPO loss on batch $\mathcal{B}$\;
    optimize $\hat{f}$ with distillation loss on batch $\mathcal{B}$\; 
    }
   }
}
\end{algorithm}

\begin{figure}[h]
\centering
\subfigure[Shooter Game ]{
\centering
\includegraphics[width=0.22\textwidth]{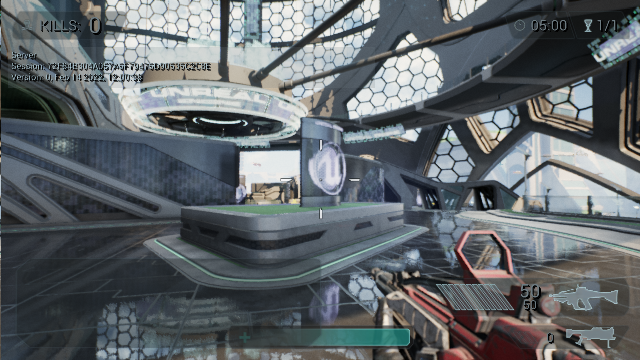}
}\quad%
\subfigure[Action RPG Game]{
\centering
\includegraphics[width=0.22\textwidth]{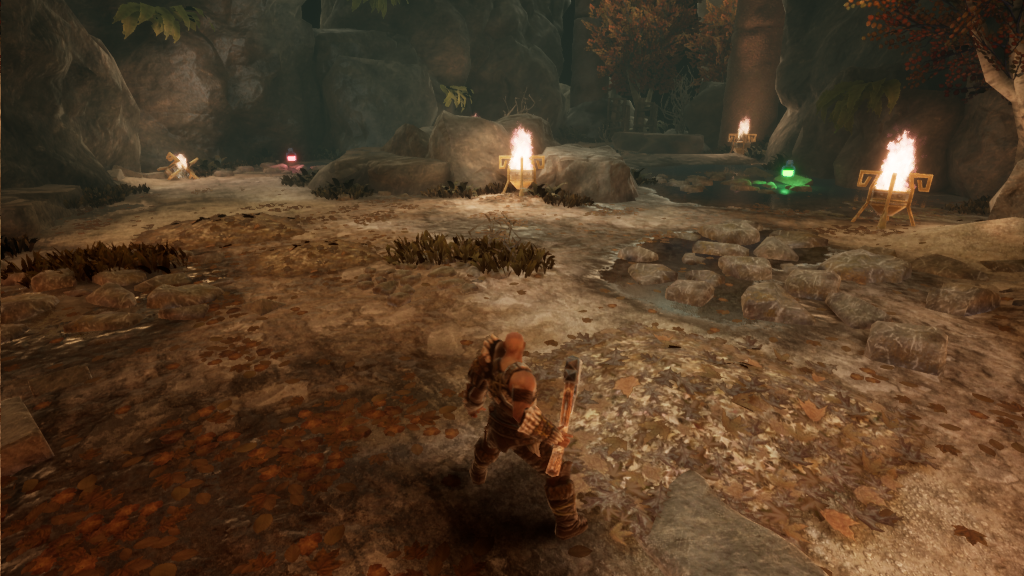}
%
}
\caption{The screenshots of the two used unreal engine video games.}
\label{fig:example_screenshot}
\end{figure}

\section{Experiments}


\begin{figure}[t]
\centering
\subfigure[$K$ = 20]{
\centering
\includegraphics[width=0.24\textwidth]{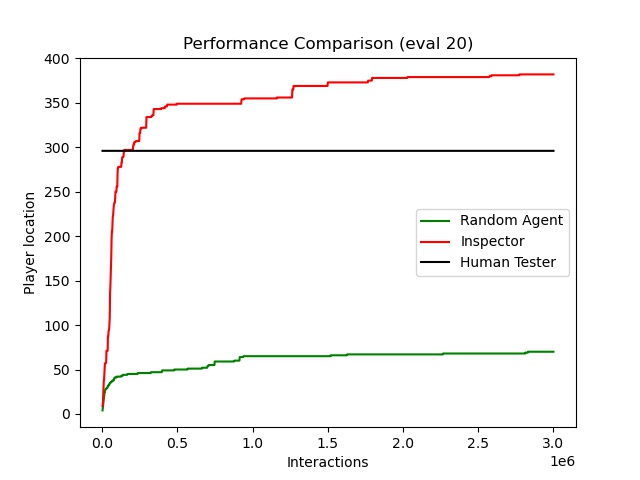}
}%
\subfigure[$K$ = 30]{
\centering
\includegraphics[width=0.24\textwidth]{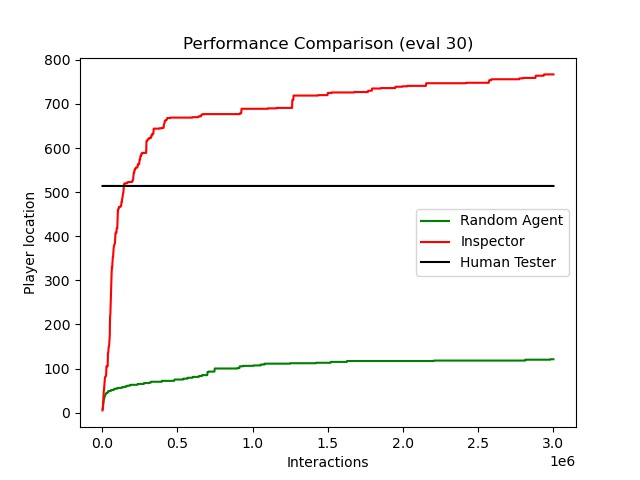}
}%
\caption{The player location coverage results in Shooter Game. $K$ represents the hyper-parameter for location discretization of the game map. 
}
\label{fig:exploration_curve_shooter_game}
\end{figure}

\begin{figure}[t]{
\centering
\subfigure[Random Agent]{
\centering
\includegraphics[width=0.16\textwidth]{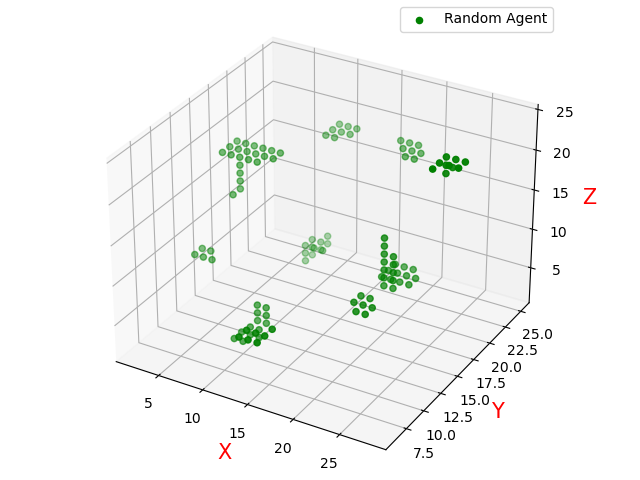}
}%
\subfigure[\underline{Inspector
}]{
\centering
\includegraphics[width=0.16\textwidth]{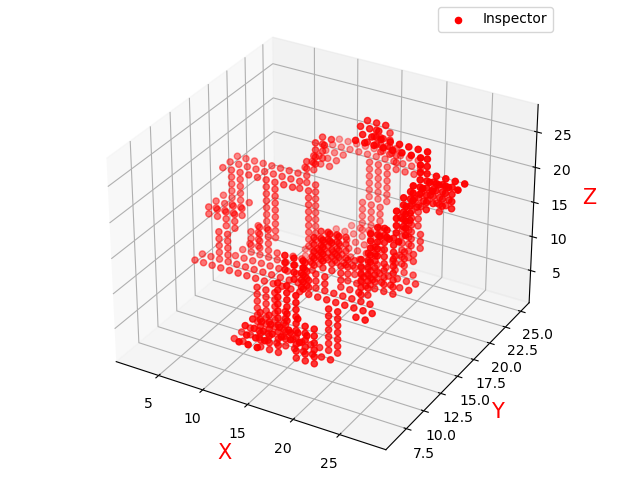}
}%
\subfigure[Human Tester]{
\centering
\includegraphics[width=0.16\textwidth]{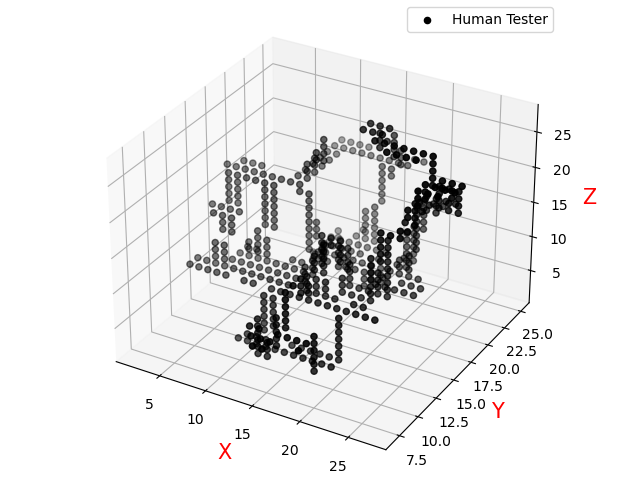}
}%

}
\caption{Visualization of the player location coverage results in Shooter Game. }
\label{fig:exploration_scatter_shooter_game}
\end{figure}

\begin{figure}[t]
\centering
\subfigure[$K$ = 20]{
\centering
\includegraphics[width=0.24\textwidth]{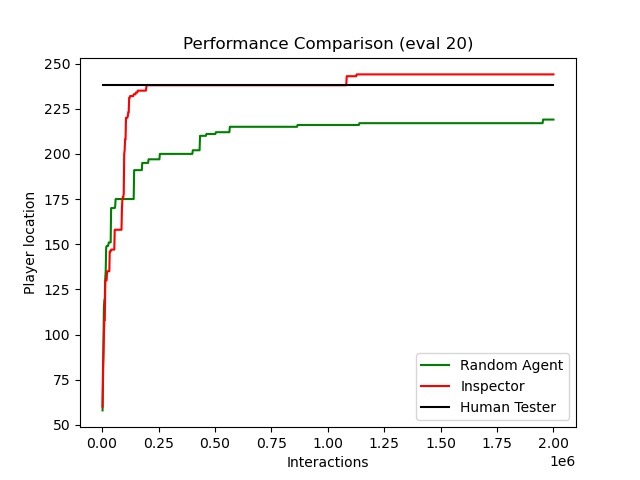}
}%
\subfigure[$K$ = 30]{
\centering
\includegraphics[width=0.24\textwidth]{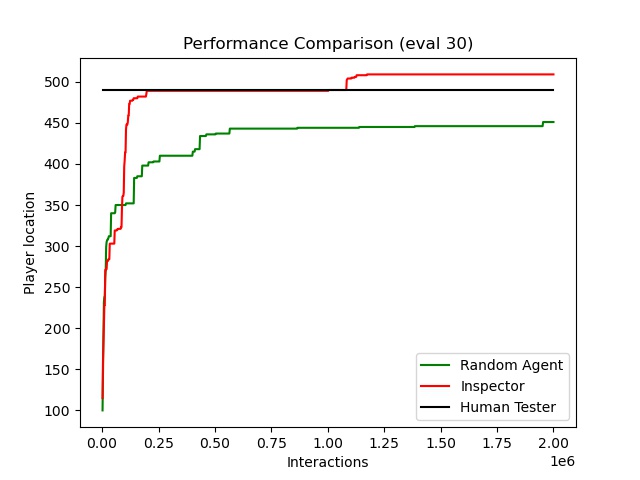}
}%
\caption{The player location coverage results in Action RPG Game. $K$ represents the hyper-parameter for location discretization of the game map.
}
\label{fig:exploration_curve_action_rpg_game}
\end{figure}
\begin{figure}[t]{
\subfigure[Random Agent]{
\centering
\includegraphics[width=0.16\textwidth]{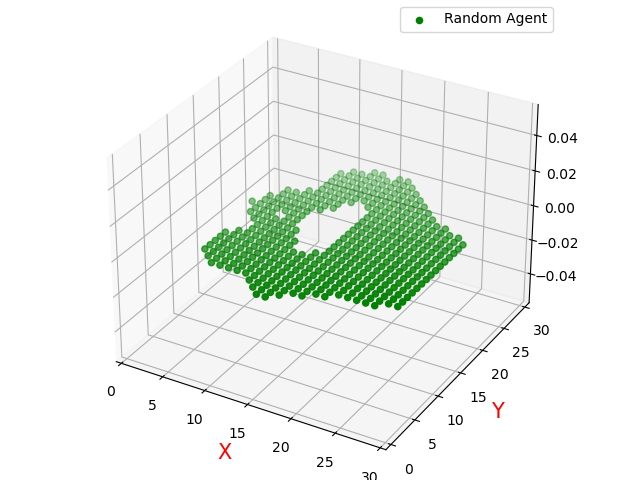}
}%
\subfigure[\underline{Inspector
}]{
\centering
\includegraphics[width=0.16\textwidth]{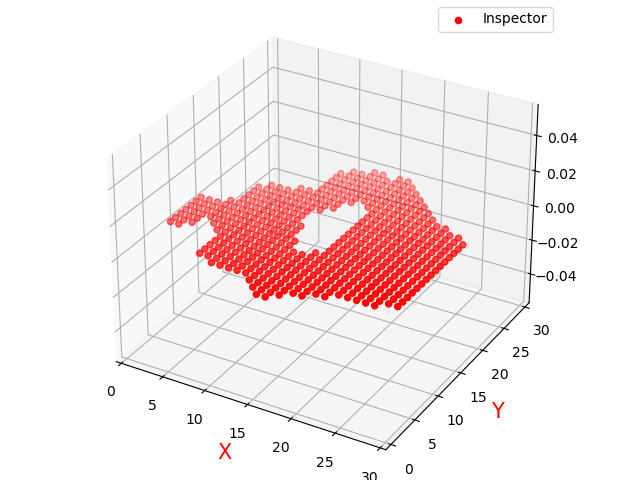}
}%
\subfigure[Human Tester]{
\centering
\includegraphics[width=0.16\textwidth]{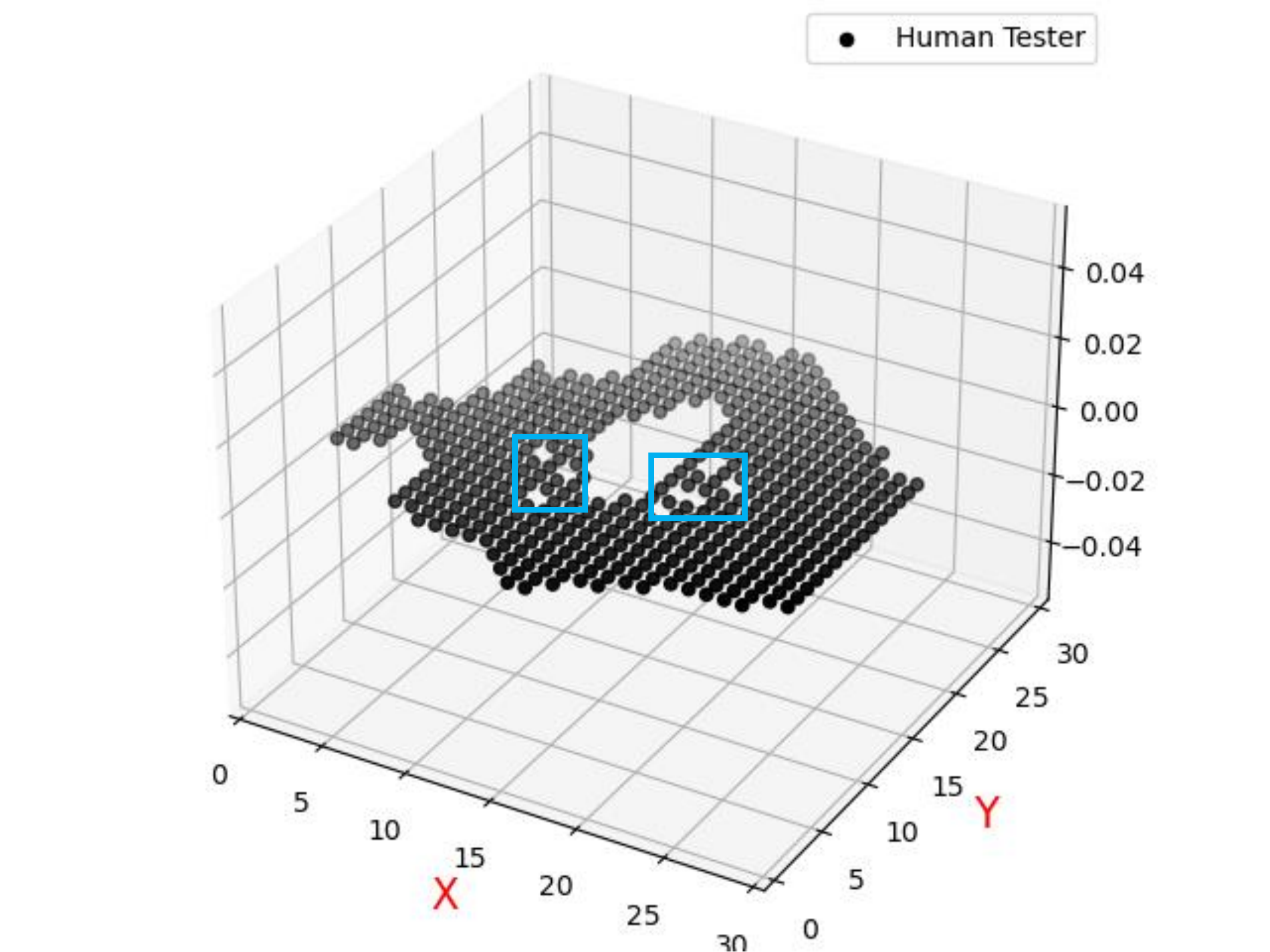}
}%

}
\caption{Visualization of the player location coverage results in Action RPG Game. }
\label{fig:exploration_scatter_action_rpg_game}
\end{figure}

To demonstrate the effectiveness of our Inspector agent, we 
apply our methods to two popular video games developed by Unreal Engine\footnote{Unreal Engine (https://www.unrealengine.com, UE in short) is a famous game engine developed by Epic Games, supporting a wide range of desktop, mobile, console and virtual reality platforms.}: Shooter \li{Game}
and Action RPG \li{Game}.
The two games belong to the two most popular game categories\footnote{Refer to https://www.microsoft.com/store/most-played/games/xbox. Most played games from Microsoft Xbox Store.}\li{, shooter and action-adventure} games, respectively. 
We first introduce the experimental setup in Section~\ref{exp:setting}, which includes game \li{descriptions} and implementation details.
After that, we show the empirical results of each module \li{in} Inspector respectively:
super-human coverage results for game space explorer (in Section~\ref{exp:exploration}), few-shot detection results for key object detector (in Section~\ref{exp:detection}), and imitation learning behaviors for human-like object investigator (in Section~\ref{exp:investigation}). 
Moreover, we also show the potential bugs \li{discovered by Inspector in these two video games} 
(in Section~\ref{exp:bugs}).   
Finally, we record the demo video of Inspector, which clearly shows \li{different stages of the whole} automated testing process
~(in Section~\ref{exp:demo_video}). 


\subsection{Experimental Setup}\label{exp:setting}

\subsubsection{Game description} 
Two popular video games, Shooter \li{Game}, and Action RPG \li{Game}, are chosen to demonstrate the effectiveness of our agent.
Shooter Game is a \li{first-person} shooter (FPS) game, while Action RPG Game is an action-adventure game. 
Both of them are complex and large \li{video} games developed by \li{Unreal Engine}. 
The screenshot sizes of the two games \li{are both} (640, 360, 3). 
In Shooter \li{Game}, the player has seven actions to take: \{Forward, Back, Left, Right, Turn left, Turn right, Jump\}. 
In Action RPG \li{Game}, the player has six actions to take: \{Forward, Back, Left, Right, Turn left, Turn right\}.
Fig.~\ref{fig:example_screenshot} depicts the example screenshots of the two games. 
We use the UnrealCV plugin\footnote{https://github.com/unrealcv/unrealcv} to facilitate the communication between the games and \li{the Inspector agent}. 

\begin{figure*}[t]
\centering
\subfigure[$0/4$ of a circle]{
\begin{minipage}[t]{0.18\linewidth}
\centering
\includegraphics[width=1.0\linewidth]{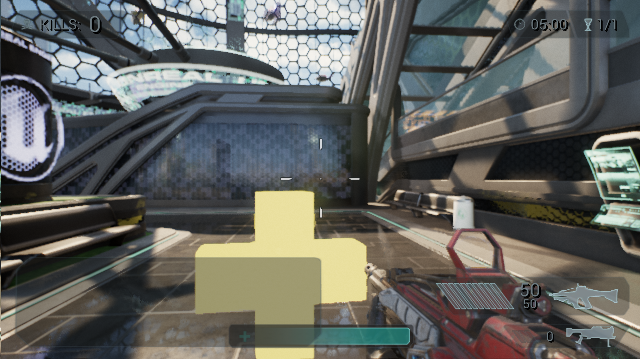}
\end{minipage}%
}%
\subfigure[$1/4$ of a circle]{
\begin{minipage}[t]{0.18\linewidth}
\centering
\includegraphics[width=1.0\linewidth]{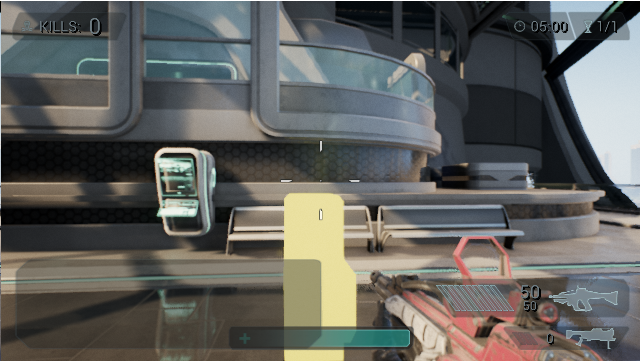}
\end{minipage}%
}%
\subfigure[$2/4$ of a circle]{
\begin{minipage}[t]{0.18\linewidth}
\centering
\includegraphics[width=1.0\linewidth]{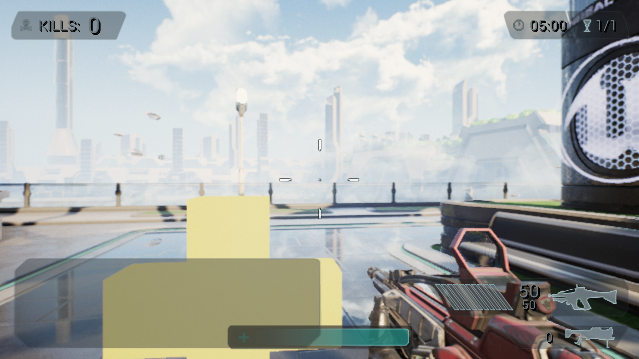}
\end{minipage}
}%
\subfigure[$3/4$ of a circle]{
\begin{minipage}[t]{0.18\linewidth}
\centering
\includegraphics[width=1.0\linewidth]{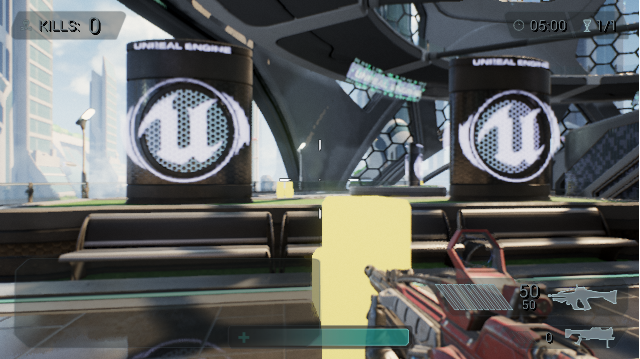}
\end{minipage}
}%
\subfigure[$4/4$ of a circle]{
\begin{minipage}[t]{0.18\linewidth}
\centering
\includegraphics[width=1.0\linewidth]{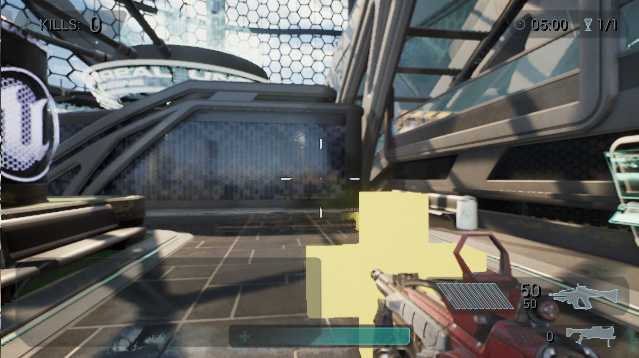}
\end{minipage}
}
\caption{Snapshots of the video showing how the learned investigation policy investigates the health pack in Shooter Game. 
}
\label{fig::imitation}
\end{figure*}

\begin{figure*}[t]
\centering
\subfigure[Exploring this area.]{
\begin{minipage}[t]{0.18\linewidth}
\centering
\includegraphics[width=1.0\linewidth]{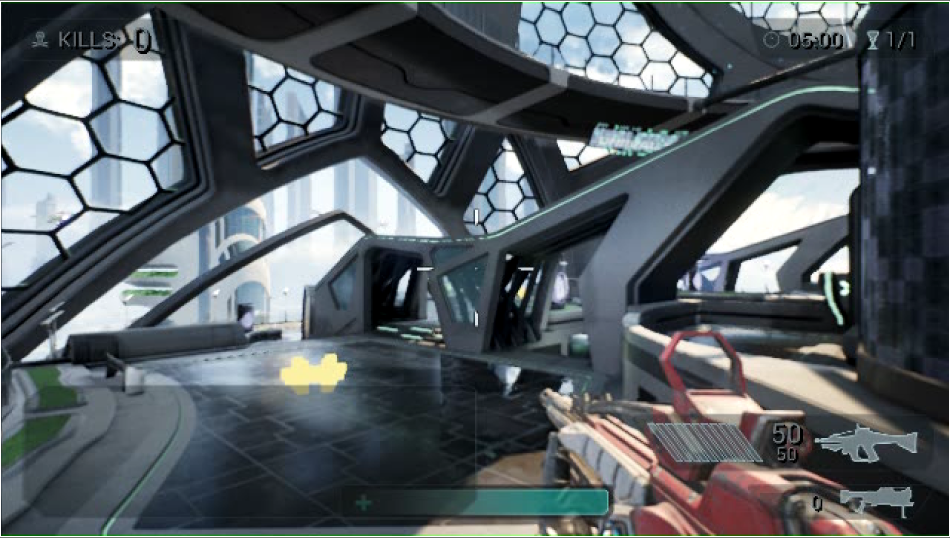}
\end{minipage}%
}\label{fig::demo_video_a}
\subfigure[Finding the health pack.]{
\begin{minipage}[t]{0.18\linewidth}
\centering
\includegraphics[width=1.0\linewidth]{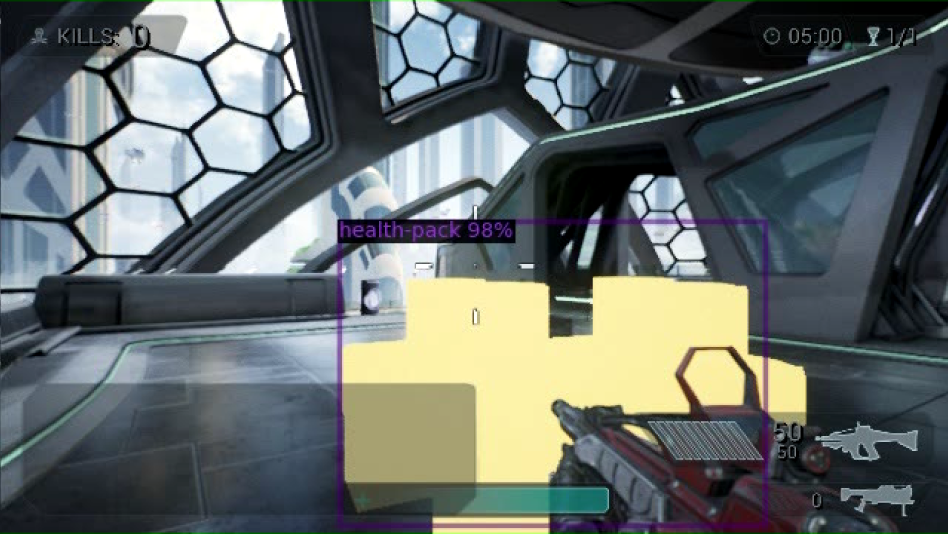}
\end{minipage}%
}%
\subfigure[Turning 180 degrees.]{
\begin{minipage}[t]{0.18\linewidth}
\centering
\includegraphics[width=1.0\linewidth]{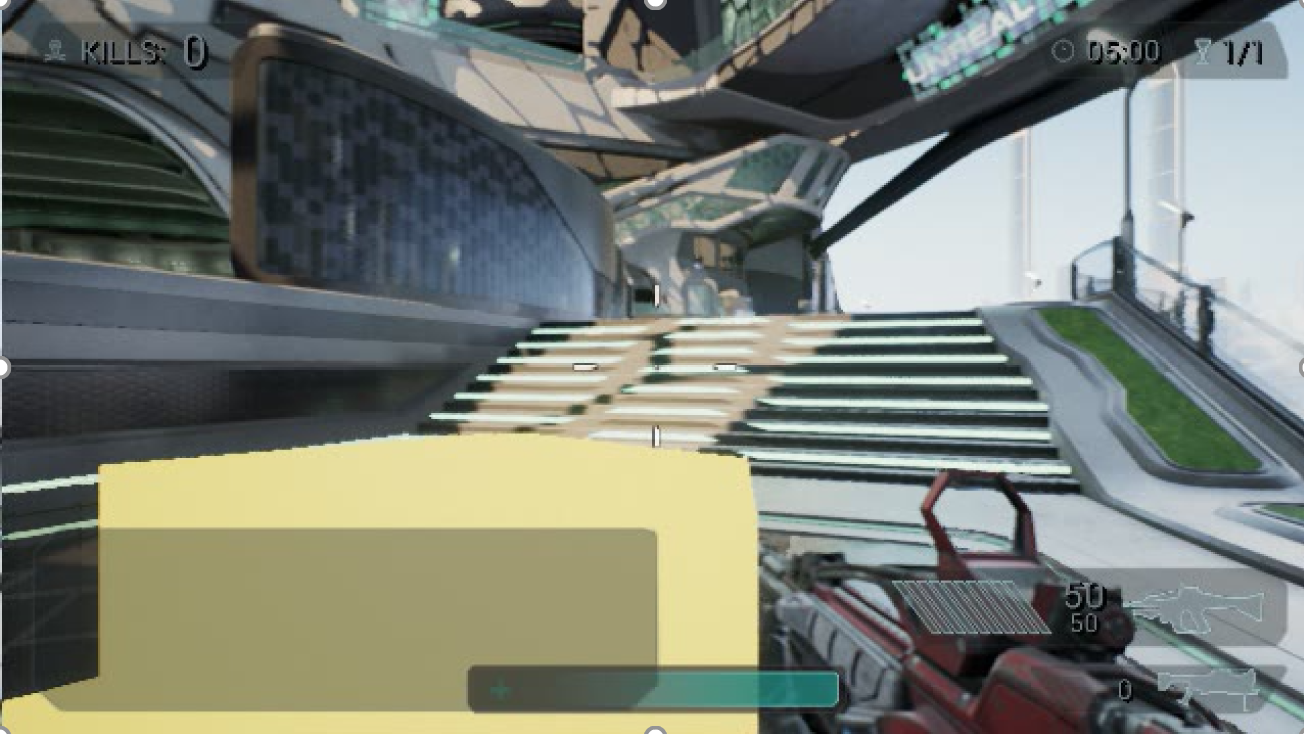}
\end{minipage}
}%
\subfigure[Turning 360 degrees.]{
\begin{minipage}[t]{0.18\linewidth}
\centering
\includegraphics[width=1.0\linewidth]{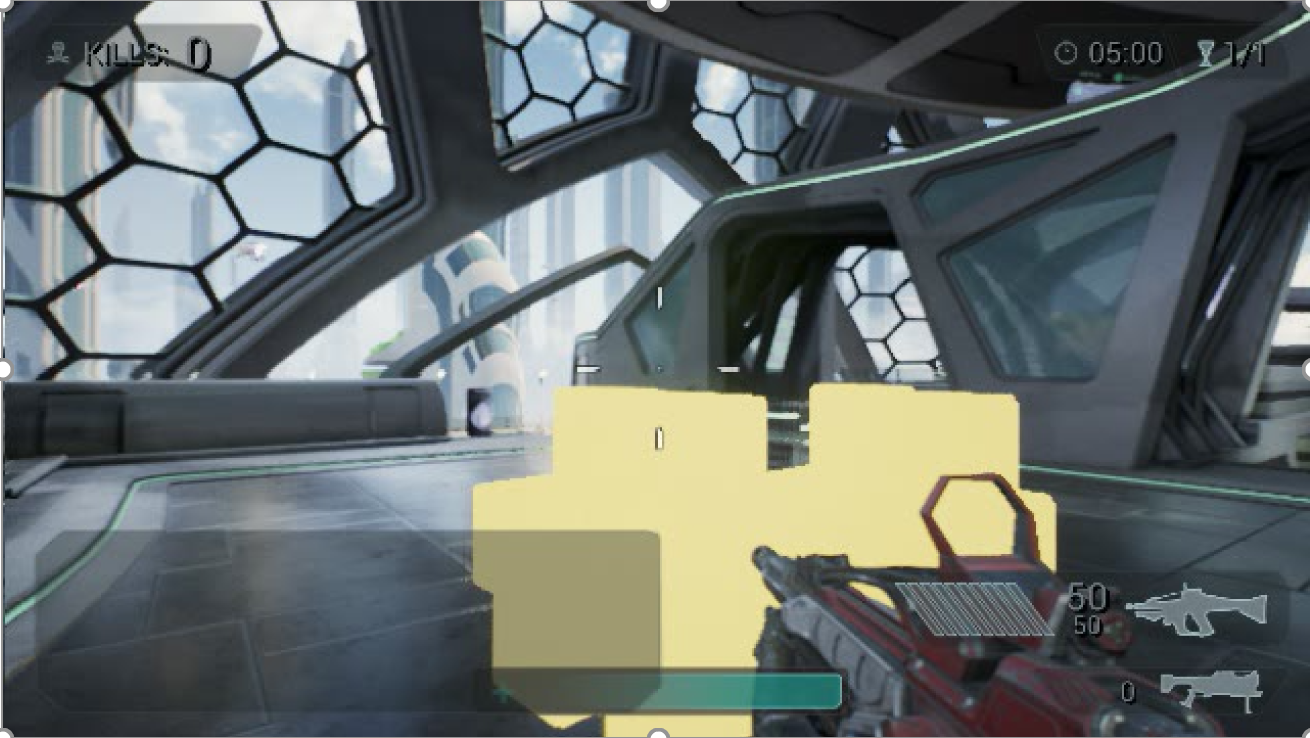}
\end{minipage}
}%
\subfigure[Re-exploring new areas.]{
\begin{minipage}[t]{0.18\linewidth}
\centering
\includegraphics[width=1.0\linewidth]{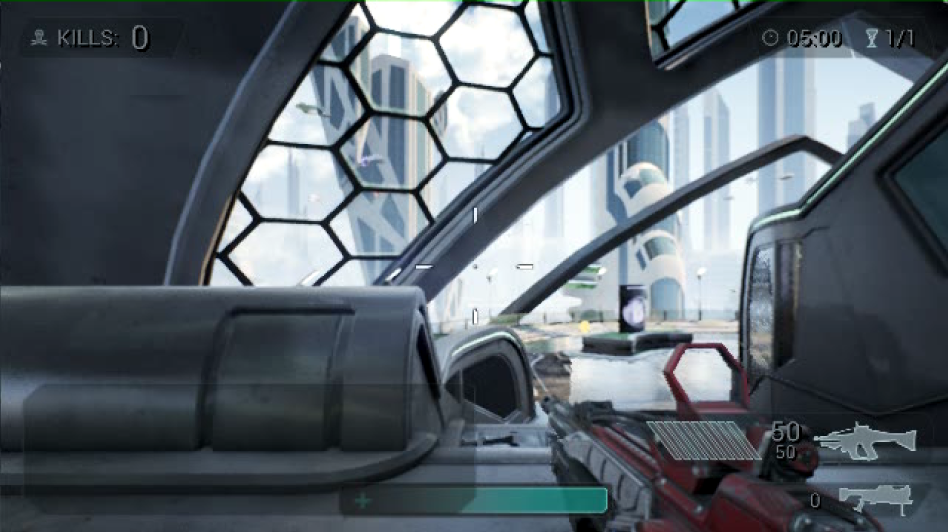}
\end{minipage}
}%
\caption{
Snapshots of the demo video for the whole testing process of Inspector in Shooter Game.  
}
\label{fig::demo_video}
\end{figure*}

\subsubsection{Implementation Details}
For the game space explorer, we use the PPO algorithm to train an exploration policy based on the \li{RND-based reward function}. The policy network consists of four convolution layers and a fully connected layer. 
Both the predictor network and the target network adopt the same network structure that comprises three convolution layers and three fully-connected layers, \li{in which} the output embedding size is 512. 
\li{During training}, the batch size for updating the parameters of the above networks is $2000$, and the number of epochs for optimization
 is 20. We use the Adam optimizer~\cite{b37} with \li{a learning rate of $3e-4$}.

For the key object detector, we use Faster R-CNN  as our base detector and Resnet-101~\cite{b35} with a Feature Pyramid Network~\cite{b38} as the backbone. 
The detection model is trained using the SGD optimizer with a mini-batch size of 16, momentum of 0.9, and weight decay of $0.0001$.  
During the base training period, the learning rate is 0.02. During the few-shot fine-tuning period, the learning rate is 0.001. 

For the human-like object investigator, the network structure comprises four convolution layers, four fully connected layers, and a softmax layer. The output represents \li{the probability of actions}.  We use the SGD optimizer with the learning rate of $0.001$, and the batch size is $256$.

For the integrated system, the threshold of the bounding box size is $18000$, while the threshold of the classification probability is $95\%$. 
The two thresholds determine when to start the investigation process for Inspector, and the length of an investigation process is 200 timesteps. 




\subsection{{Super-human Coverage Results }}\label{exp:exploration}

\li{In this subsection, we evaluate the ability of Inspector to navigate and explore the whole map, and show that Inspector achieves super-human map coverage in both games.}
Since the player location is a \li{3-dimensional} continuous vector, 
\li{we discretize the whole game map with a volume  of $K^{3}$ to compute the \li{map} coverage result,}
where $K$ is the hyper-parameter for 
discretization.  

Fig.~\ref{fig:exploration_curve_shooter_game}, Fig.~\ref{fig:exploration_curve_action_rpg_game} show the coverage results of 
Random agent, 
Inspector, 
and Human tester in Shooter Game and Action RPG Game, respectively.
As the training proceeds,
Inspector (red) 
keeps exploring the new areas, and outperforms the results of Human tester (black) and Random agent (green), in both games. 

Fig.~\ref{fig:exploration_scatter_shooter_game}, Fig.~\ref{fig:exploration_scatter_action_rpg_game} also depict the 3D scatter plot of the explored regions by those three methods 
($K=30$).  
From the results \li{on} Shooter \li{Game}, 
we can see that Random agent (green) 
\li{only covers a small region around the random starting player location.}
Although Human tester (black) can roughly cover the game map, 
there are \li{still} a few places that remain unexplored (such as some places around the center of the map).  
In contrast, Inspector (red) can cover the whole game space thoroughly. 
From the results \li{on} Action RPG \li{Game}, we can see that Random agent (green) can achieve good results in this map, since the map size of Action RPG Game is much smaller than that of Shooter Game. 
However, Random agent (green) \li{cannot} 
cover some regions on the edge of the map. 
Similar to the \li{results} 
in Shooter Game, Human tester (black) can roughly cover the map but still have some places unexplored (such as those blank areas in Fig.~\ref{fig:exploration_scatter_action_rpg_game} (c)).
\li{In comparison, }
Inspector (red) can consistently achieve the full coverage of the game map. 

\subsection{Few-shot Detection Results}\label{exp:detection}


In this subsection, \li{we evaluate the key object detector on Shooter Game.}
\li{We collect 20 examples of the health pack for few-shot training, and 9 examples for testing. }



\begin{figure}[h]
\centering
\subfigure[]{
\centering
\includegraphics[width=0.31\linewidth]{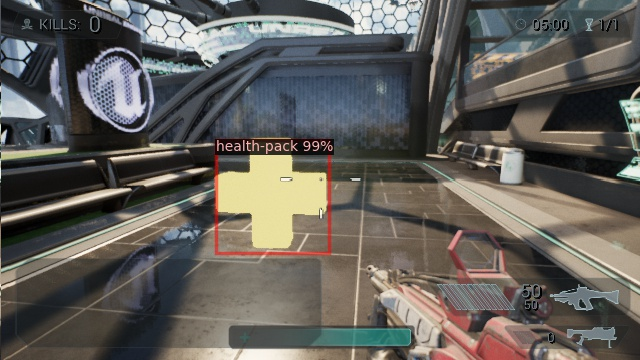}
}%
\subfigure[]{
\centering
\includegraphics[width=0.31\linewidth]{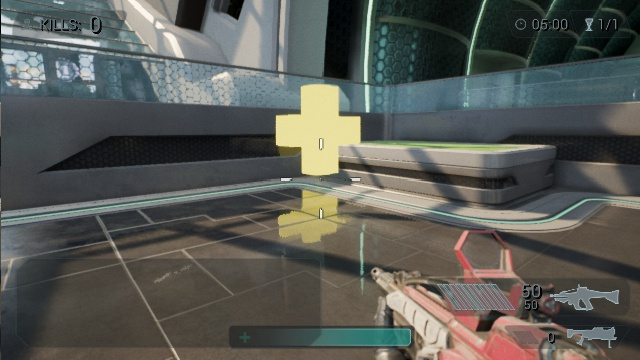}
}%
\subfigure[]{
\centering
\includegraphics[width=0.31\linewidth]{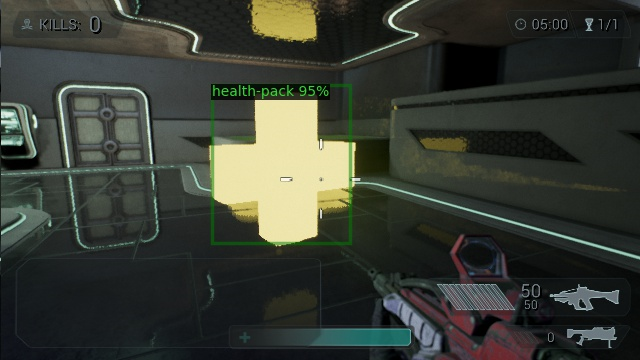}
}%

\subfigure[]{
\centering
\includegraphics[width=0.31\linewidth]{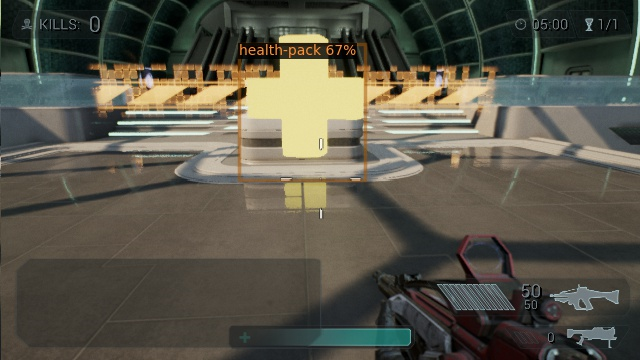}
}%
\subfigure[]{
\centering
\includegraphics[width=0.31\linewidth]{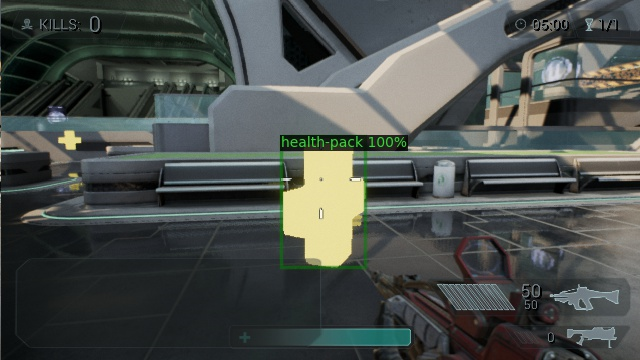}
}%
\subfigure[]{
\centering
\includegraphics[width=0.31\linewidth]{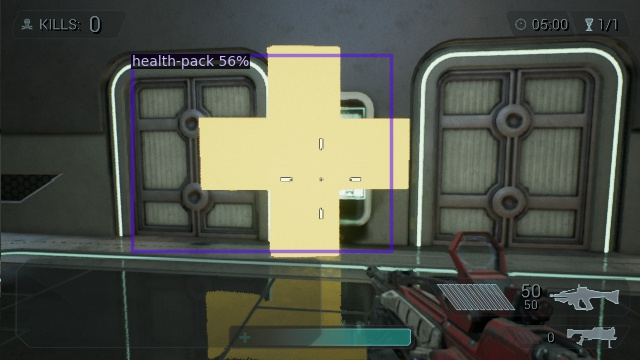}
}%

\subfigure[]{
\centering
\includegraphics[width=0.31\linewidth]{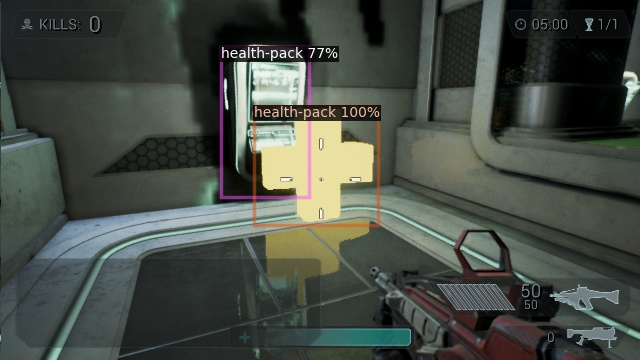}
}%
\subfigure[]{
\centering
\includegraphics[width=0.31\linewidth]{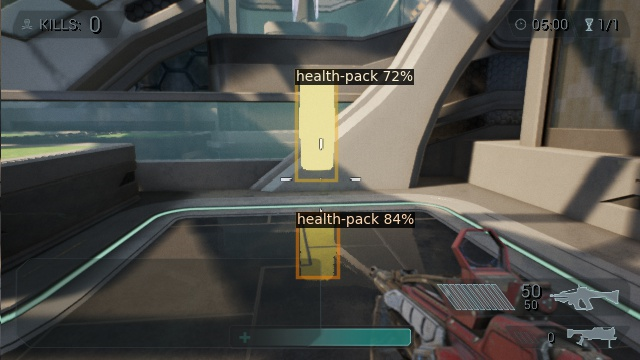}
}%
\subfigure[]{
\centering
\includegraphics[width=0.31\linewidth]{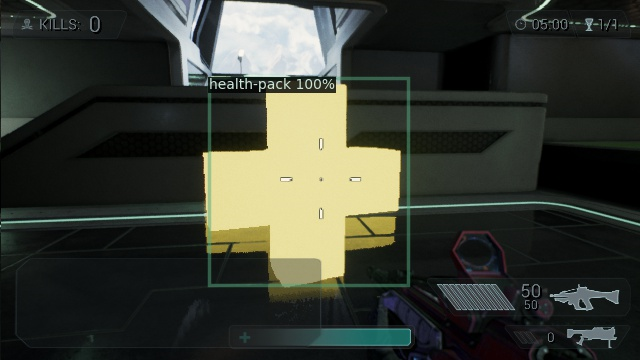}
}%
\caption{The detection results of the key object detector in Shooter Game.}
\label{fig::fsdet}
\end{figure}

Fig.~\ref{fig::fsdet} shows the detection results of the learned key object detector on the \li{test examples}. We can see that 
\li{the key object detector accurately detects the health pack in most cases (7 out of 9 examples). }
It is worth mentioning that all the backgrounds in 9 test examples are never seen in the 20 training examples, \li{which demonstrates the effectiveness of the few-shot object detection.} 

\begin{figure*}[t]
\centering
\subfigure[An unimpressive corner in the Shooter Game map.]{
\centering
\includegraphics[width=0.3\textwidth]{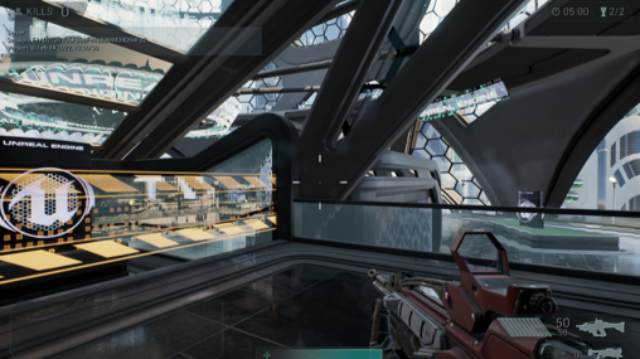}
}\quad\quad
\subfigure[Bug: the player stands without support under the feet.]{
\centering
\includegraphics[width=0.3\textwidth]{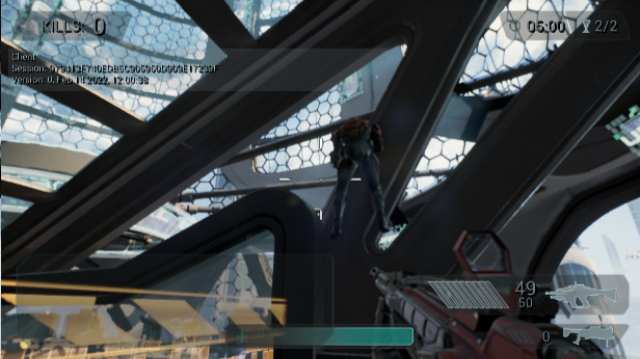}
}%
\caption{The standing bug discovered by Inspector in Shooter Game.}
\label{fig:bug_shooter_game}
\end{figure*}



\begin{figure*}[t]
\centering
\subfigure[Close to a rock in the Action RPG Game map.]{
\centering
\includegraphics[width=0.3\textwidth]{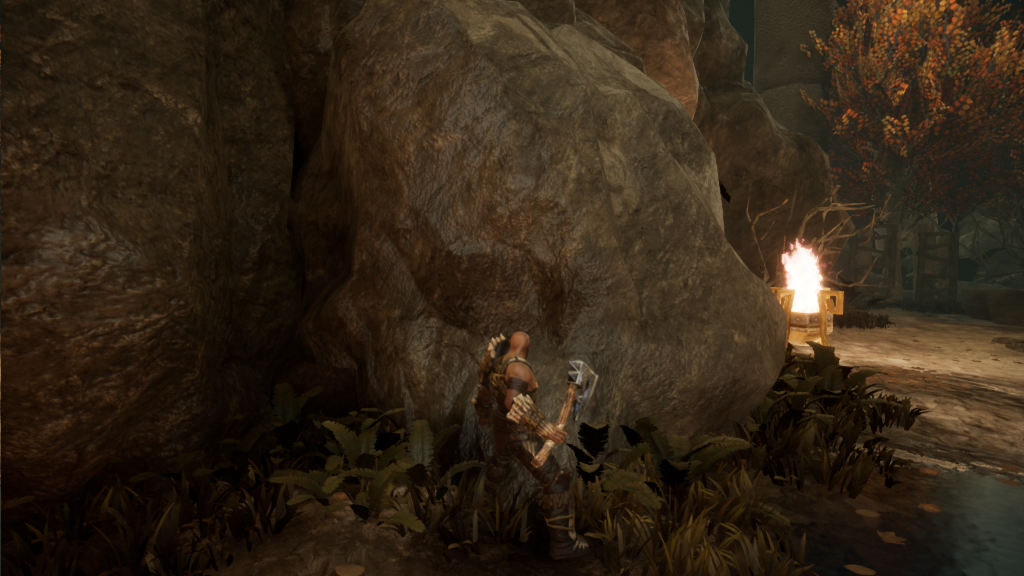}
}\quad\quad
\subfigure[Bug: the player clips into the rock.]{
\centering
\includegraphics[width=0.3\textwidth]{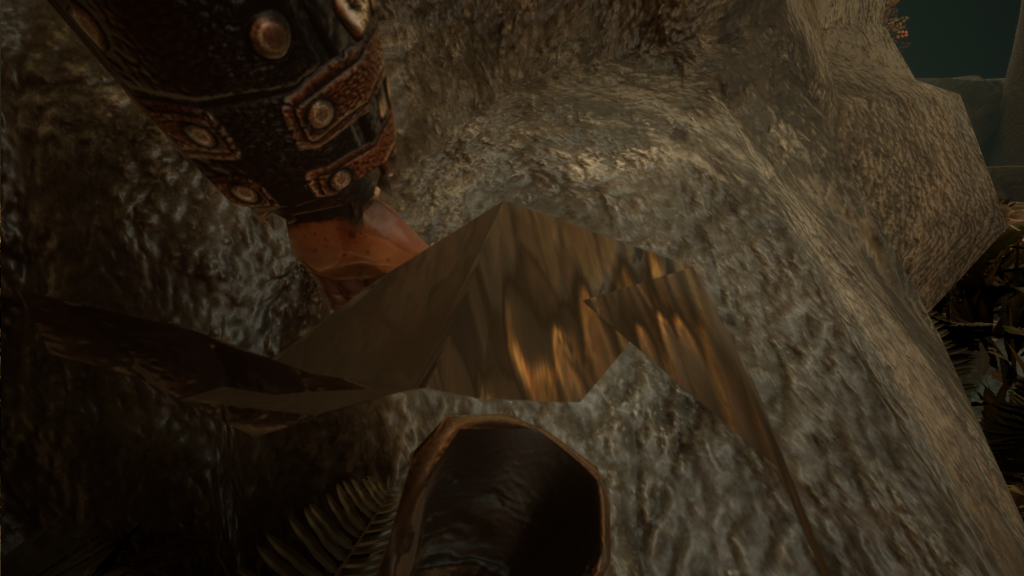}
}%
\caption{The collision bug discovered by Inspector in Action RPG Game.}
\label{fig:bug_action_rpg}
\end{figure*}



\subsection{{Imitation Learning Behaviors}}\label{exp:investigation}

In this subsection, we show the learned behaviors by pixel-based imitation learning for human-like object investigation in Shooter Game. 
We first collect 25 trajectories (about 5000 labeled \li{screenshots and action pairs} in total) from human testers for investigating the health pack. 
\li{To investigate health packs, human testers circle around the health pack to make sure it looks correct from every angle.}
After using these human demonstrations to learn the investigation policy, we evaluate the performance of the investigation policy by deploying it to interact with the health pack within the game. 
We record the video of this process in \url{https://github.com/Inspector-GameTesting/Inspector-GameTesting/imitation_learning_behavior.avi}. The snapshots in Fig.~\ref{fig::imitation} show the circular motion of the learned investigation agent in 0, 90, 180, 270, and 360 degrees, respectively. 
The result shows that the 
\li{human-like investigation behaviors} can be learned by the pixel-based imitation learning method with limited demonstrations.



\subsection{{Potential Bugs Discovered by Inspector}}\label{exp:bugs}

In this subsection, we show two potential bugs hidden in Shooter Game and Action RPG Game, \li{which are} discovered by Inspector during the exploration stage. 
The two discovered bugs are shown in Fig.~\ref{fig:bug_shooter_game} and Fig.~\ref{fig:bug_action_rpg}, respectively. 
To be specific,  Fig.~\ref{fig:bug_shooter_game} (a) shows us an unimpressive corner in the Shooter Game map, and Inspector successfully finds a bug where the player can stand without support under the feet after jumping in this corner.  
Fig.~\ref{fig:bug_action_rpg} shows that when close to a specific rock in the Action RPG Game map, Inspector takes a step forward, and then clips into this rock. 
We can see that these two bugs are concealed and might be not so easy for the human testers to find. 
The two found bugs suggest that Inspector has the strong ability to explore the game space and find the potential bugs in \li{video games}.

\subsection{{Demo Videos of Inspector}}\label{exp:demo_video}

To better show the capabilities of the Inspector agent for automated game testing, 
we record several demo videos for the whole testing progress of the agent, which are available at \url{https://github.com/Inspector-GameTesting/Inspector-GameTesting/}.
Fig.~\ref{fig::demo_video} shows the key snapshots in one of the demo videos: At the beginning, the agent explores this area via curiosity (a).  
After some time, the  agent detects the health pack during the exploration (b). 
Then, the  agent performs the human-like investigation to the health pack, by making a full circle around it (c)(d).
When the investigation is done, the  agent keeps exploring new areas, 
\li{until it finds} the next health pack to investigate (e).

\section{Conclusion and Future Work}

In this work, we built a general pixel-based automated game testing agent/tool, named Inspector, consisting of three key modules: a game space explorer, a key object detector, and a human-like object investigator. Inspector has two main advantages over previous methods/agents: its larger application scope without the limitation of accessing game source code, and its ability to discover hidden and difficult bugs through human-like investigations with key objects. 

For future work, there are multiple directions to advance Inspector. 
First, we only tested it with two games both on the Unreal engine in this work. We will test it over other game engines (e.g., Unity) and across different devices (e.g., PC, Console, etc) with only the screenshot input.
Second, it is worthy of investigation on how to enable Inspector to explore more complex games, e.g., a game with  multi-room navigation, in which the agent needs to first  pick up a key somewhere in a room, and then  open another  room with the key to enter the room.
Third,  the key object detector in this work handles each kind of object with a separate detection model and needs human-labeled data for each kind of object. We will study how to detect multiple kinds of objects with a single model, e.g., through multi-task learning, so as to further reduce human labeling costs.
Fourth, human players/testers may take different actions for different kinds of objects. Similarly, we will train a single model to perform  human-like investigations over multiple kinds of objects. Last but not least, we did not consider the bug detection module from screenshots in this work. We will build a complete game testing service by enhancing Inspector with a \li{pixel-based} bug detection module, e.g.,\cite{b40}.


\begin{thebibliography}{00}





\bibitem{b1} D. Lin, C. Bezemer and A. E. Hassan, “Studying the urgent updates of popular games on the steam platform," \textit{Empirical Software Engineering}, vol. 22, pp.2095-2126, 2017.

\bibitem{b2} C. Politowski, F. Petrillo, Y. Gäel Guéhéneuc, “A Survey of Video Game Testing,” \textit{CoRR}, vol. abs/2103.06431, https://arxiv.org/abs/2103.06431. 2021.

\bibitem{b3} S. Aleem, L. F. Capretz and F. Ahmed, “Critical Success Factors to Improve the Game Development Process from a Developers Perspective," \textit{Journal of Computer Science and Technology}, vol. 31, pp.925-950, 2016.

\bibitem{b4} R. E. S. Santos, C. V. C. de Magalhães, L. F. Capretz, J. S. C. Neto, F. Q. B. da Silva, and A. Saher, “Computer games
are serious business and so is their quality: Particularities of
software testing in game development from the perspective of
practitioners,” \textit{CoRR}, vol. abs/1812.05164, https://arxiv.org/abs/1812.05164, 2018.

\bibitem{b5} C. Cho, D. Lee, K. Sohn, C. Park, and J. Kang, “Scenario-based approach for blackbox load testing of online game servers,” in \textit{International Conference on Cyber-Enabled Distributed Computing and Knowledge Discovery}, pp. 259–265, 2010.

\bibitem{b6} M. Ostrowski and S. Aroudj, “Automated regression testing
within video game development,” \textit{GSTF Journal on Computing
(JoC)}, vol. 3, no. 2, pp. 1–5, 2013.

\bibitem{b7} S. Iftikhar, M. Z. Iqbal, M. U. Khan and W. Mahmood, "An automated model based testing approach for platform games," \textit{ACM/IEEE 18th International Conference on Model Driven Engineering Languages and Systems (MODELS)}, pp. 426-435, 2015. 

\bibitem{b8} J. Hernández Bécares, L. Costero, and P. Gómez-Martín, “An approach to automated videogame beta testing,” \textit{Entertainment
Computing}, vol. 18, 2016.

\bibitem{b9} C. Holmgard, M. C. Green, A. Liapis, and J. Togelius, “Automated playtesting with procedural personas through mcts with evolved heuristics,” \textit{IEEE Transactions on Games}, vol. 11, no. 4, pp. 352–362, 2018.

\bibitem{b10} S. Ariyurek, A. Betin-Can and E. Surer, "Automated Video Game Testing Using Synthetic and Humanlike Agents," in \textit{IEEE Transactions on Games}, vol. 13, no. 1, pp. 50-67, 2021.

\bibitem{b11} S. Ariyurek, A. Betin-Can and E. Surer, "Enhancing the Monte Carlo Tree Search Algorithm for Video Game Testing," in \textit{IEEE Conference on Games (CoG)}, pp. 25-32, 2020.

\bibitem{b12} L. Mugrai, F. de Mesentier Silva, C. Holmgard, and J. Togelius, “Automated playtesting of matching tile games,” in \textit{IEEE Conference on Games}, 2019.

\bibitem{b13} A. Gustavo, R. Geber, S. Hugo, C. Vincent, "Automatic computer game balancing: a reinforcement learning approach," 1111-1112. 10.1145/1082473.1082648, 2005.


\bibitem{b14} N. Ashey, P. Aline, C. Esteban. "Towards Adaptive Deep Reinforcement Game Balancing," pp. 693-700. 10.5220/0007395406930700, 2019. 

\bibitem{b15} C. Guerrero-Romero, S. M. Lucas, and D. Perez-Liebana, “Using a team of general ai algorithms to assist game design and testing,” in \textit{IEEE Conference on Computational Intelligence and Games (CIG)}. pp. 1–8, 2018.

\bibitem{b16} Y. Zhao, I. Borovikov, F. D. M. Silva, A. Beirami, J. Rupert, C. Somers, J. Harder, J. Kolen, J. Pinto, R. Pourabolghasem, J. Pestrak. H. Chaput, M. Sardari, L. Lin, S. Narravula, N. Aghdaie and K. Zaman, “Winning Is Not Everything: Enhancing Game Development With Intelligent Agents,” \textit{IEEE Transactions on Games}, 2020. 


\bibitem{b17} L. Gisslen, A. Eakins, C. Gordillo, J. Bergdahl and K. Tollmar, “Adversarial Reinforcement Learning for Procedural Content Generation," \textit{IEEE Conference on Games}, 2021.


\bibitem{b18} J. Pfau, J. Smeddinck, and R. Malaka, “Automated game testing
with icarus: Intelligent completion of adventure riddles via
unsupervised solving,” in \textit{CHI PLAY}, 2017.

\bibitem{b19} Z. Zhan, B. Aytemiz, and A. M. Smith, “Taking the scenic route: Automatic exploration for videogames,” in \textit{KEG@AAAI, ser. CEUR Workshop Proceedings}, vol. 2313. CEUR-WS.org,  pp. 26–34, 2019.

\bibitem{b20} J. Bergdahl, C. Gordillo, K. Tollmar, and L. Gisslen, “Augmenting automated game testing with deep reinforcement learning,” in \textit{IEEE Conference on Games (CoG)}, pp. 600–603, 2020.

\bibitem{b21} Y. Zheng, X. Xie, T. Su, L. Ma, J. Hao, Z. Meng, Y. Liu, R. Shen, Y. Chen, and C. Fan, “Wuji: Automatic online combat game testing using evolutionary deep reinforcement learning,” in Proceedings of the 34th IEEE/ACM International Conference on Automated Software Engineering. pp. 772–784, 2019.

\bibitem{b22} C. Gordillo, J. Bergdahl, K. Tollmar and L. Gisslen. “Improving Playtesting Coverage via Curiosity Driven Reinforcement Learning Agents,” in \textit{IEEE
Conference on Games (CoG)}, 2021.

\bibitem{b23} K. Chang, B. Aytemiz, and A. M. Smith, “Reveal-more: Amplifying human effort in quality assurance testing using automated exploration,” in \textit{IEEE Conference on Games (CoG)}, pp. 1–8, 2019.

\bibitem{b24} A. Aubret, L. Matignon, and S. Hassas, “A survey on intrinsic motivation
in reinforcement learning,” \textit{CoRR}, vol. abs/1908.06976, https://arxiv.org/abs/1908.06976, 2019.


\bibitem{b25} M. Bellemare, S. Srinivasan, G. Ostrovski, T. Schaul, D. Saxton, and
R. Munos, “Unifying count-based exploration and intrinsic motivation,”
in \textit{Advances in Neural Information Processing Systems}, vol. 29. Curran Associates, Inc., pp. 1471–1479,  2016.


\bibitem{b26} R. S. Sutton and A. G. Barto, ``Reinforcement Learning: An Introduction". Cambridge, MA, USA: MIT Press, 2018.

\bibitem{b27} K. Shao, Z. Tang, Y. Zhu, N. Li and D. Zhao, ``A Survey of Deep Reinforcement Learning in Video Games", \textit{CoRR}, vol. abs/1912.10944,  http://arxiv.org/abs/1912.10944, 2019.


\bibitem{b28} R. R. Torrado, P. Bontrager, J. Togelius, J. Liu and D. Perez-Liebana, "Deep Reinforcement Learning for General Video Game AI," 2018 \textit{IEEE Conference on Computational Intelligence and Games (CIG)}, pp. 1-8, doi: 10.1109/CIG.2018.8490422, 2018.

\bibitem{b29} Y. Burda, H. Edwards, A. Storkey and O. Klimov, ``Exploration by random network distillation". In \textit{International Conference on Learning Representations (ICLR)}, 2019.


\bibitem{b30} J. Schulman, F. Wolski, P. Dhariwal, A. Radford, and O. Klimov, “Proximal policy optimization algorithms,” \textit{CoRR}, vol. abs/1707.06347, https://arxiv.org/abs/1707.06347, 2017.

\bibitem{b31} M. Sun, V. Kurin, G. Liu, S. Devlin, T. Qin, K. Hofmann, and S. Whiteson, “You May Not Need Ratio Clipping in PPO,” \textit{CoRR}, vol. abs/2202.00079, https://arxiv.org/abs/2202.00079, 2022.

\bibitem{b32} R. Liang, Y. Zhu, Z. Tang, M. Yang and X. Zhu, "Proximal Policy Optimization with Elo-based Opponent Selection and Combination with Enhanced Rolling Horizon Evolution Algorithm,"  \textit{IEEE Conference on Games (CoG)}, 2021, pp. 1-4, doi: 10.1109/CoG52621.2021.9619146, 2021.

\bibitem{b33} X. Wang, T. E. Huang, T. Darrell, J. E. Gonzalez and F. Yu, ``Frustratingly Simple Few-Shot Object Detection," in \textit{International Conference on Machine Learning (ICML)}, 2020.

\bibitem{b34} S. Ren, K. He, R. Girshick and J. Sun, ``Faster R-CNN: Towards Real-Time Object Detection with Region Proposal Networks," \textit{Advances in Neural Information Processing Systems}, vol. 28, 2015.

\bibitem{b35} K. He, X. Zhang, S. Ren and J. Sun, "Deep Residual Learning for Image Recognition,"  \textit{IEEE Conference on Computer Vision and Pattern Recognition (CVPR)}, pp. 770-778, 2016.


\bibitem{b36} D. A. Pomerleau. ``Efficient training of artificial neural networks for autonomous navigation," in \textit{Neural Computation}, 3(1):88–97, 1991.

\bibitem{b37} D. P. Kingma and J. Ba, ``Adam: A Method for Stochastic Optimization," In \textit{3rd International Conference for Learning Representations}, San Diego, 2015.

\bibitem{b38} T. -Y. Lin, P. Dollár, R. Girshick, K. He, B. Hariharan and S. Belongie, "Feature Pyramid Networks for Object Detection," \textit{IEEE Conference on Computer Vision and Pattern Recognition (CVPR)}, pp. 936-944, doi: 10.1109/CVPR.2017.106, 2017.

\bibitem{b39} Website of Mojang Studios. https://bugs.mojang.com/browse/MC-16731.

\bibitem{b40}  C. Ling,  K. Tollmar and L. Gisslén, ``Using Deep Convolutional Neural Networks to Detect Rendered Glitches in Video Games," \textit{Proceedings of the AAAI Conference on Artificial Intelligence and Interactive Digital Entertainment}, 16(1), 66-73, 2020.

\bibitem{b41} C. B. Browne et al., "A Survey of Monte Carlo Tree Search Methods," in \textit{IEEE Transactions on Computational Intelligence and AI in Games}, vol. 4, no. 1, pp. 1-43, 2012.



\end{thebibliography}
\end{document}